\documentclass[journal]{IEEEtran}
\ifCLASSINFOpdf
\else
\fi
\usepackage{amsfonts}
\usepackage{cite}
\usepackage{times}
\usepackage{epsfig}
\usepackage{graphicx}
\usepackage{amsmath}
\usepackage{amssymb}
\usepackage{multirow}
\usepackage{epstopdf}
\usepackage{subfigure}
\usepackage{diagbox}
\usepackage{url}
\usepackage{booktabs}
\usepackage{color}
\begin{document}
%
\title{Video Summarization with Attention-Based \\Encoder-Decoder Networks }
%
%
%

\author{Zhong~Ji\textsuperscript{1}, Kailin~Xiong\textsuperscript{1}, Yanwei~Pang\textsuperscript{1}, and Xuelong~Li\textsuperscript{2} \\ \textsuperscript{1}Tianjin University \\\textsuperscript{2}Xi'an Institute of Optics and Precision Mechanics}
\maketitle

\begin{abstract}
 This paper addresses the problem of supervised video summarization by formulating it as a sequence-to-sequence learning problem, where the input is a sequence of original video frames, the output is a keyshot sequence. Our key idea is to learn a deep summarization network with attention mechanism to mimic the way of selecting the keyshots of human. To this end, we propose a novel video summarization framework named \emph{Attentive encoder-decoder networks for Video Summarization} (AVS), in which the encoder uses a \emph{Bidirectional Long Short-Term Memory} (BiLSTM) to encode the contextual information among the input video frames. As for the decoder, two attention-based LSTM networks are explored by using additive and multiplicative objective functions, respectively. Extensive experiments are conducted on two video summarization benchmark datasets, i.e., SumMe, and TVSum. The results demonstrate the superiority of the proposed AVS-based approaches against the state-of-the-art approaches, with remarkable improvements from 0.8\% to 3\% on two datasets, respectively.
\end{abstract}


%
\IEEEpeerreviewmaketitle

\section{Introduction}
%
%
%
%
\IEEEPARstart{V}{ideo}  is inundating the Internet social platform. There are more than 300 hours¡¯ video upload per minute to YouTube. It is awfully time-consuming to browse these videos. According to Cisco¡¯s 2015 Visual Networking Index, it will take over 500 million years to watch all videos uploaded to Internet per month in the year of 2020! It is therefore becoming increasingly important to efficiently browse, manage, and retrieve these videos.

Video summarization is one of the promising techniques to address this challenge\cite{Truong2007,Money2008,Ma2002,Wang2012,Zhang2017,Xu2017}. Its goal is to produce a compact yet comprehensive summary to enable an efficient browsing experience. An ideal video summarization is that can provide users the maximum information of the target video with the shortest time. It is also useful for many other practical applications, such as video indexing\cite{hong2017}, video retrieval\cite{yanget2017}, and event detection\cite{Feng2017}.

Generally, there are two types of video summarization: storyboard and video skim. Specifically, a storyboard is based on a set of keyframes, and a video skim is composed of a number of representative video segments, called keyshots. In this work, we focus on video skim. However, it can be easily converted to the form of storyboard by selecting one or several keyframes from each keyshot.

Video summarization has been studied over two decades\cite{Truong2007,Money2008,Ma2002,Wang2012,Zhang2017,Xu2017}. During these years, many approaches have been developed by exploring cues ranging from low-level visual inconsistency\cite{Ngo2005}\cite{Cernekova2006}, attention \cite{Ma2002}\cite{Zhang2017}\cite{Ejaz2013}, to high-level semantic change of concepts \cite{Xu2017}\cite{Khosla2013} and entities in videos \cite{Lee2012}\cite{Mitra2017}. However, most of these studies focus on unsupervised leaning technique. Recently, the research focus has been extending to supervised learning approaches \cite{Gong2014,Gygli2015,Gygli2014,zhang2016,Mahasseni2017,Li2017}, which aims at explicitly learning the summarizing capability from the human labels. Usually, supervised approaches have better performance than unsupervised ones.

Among the previous supervised approaches, studies in \cite{zhang2016} and \cite{Mahasseni2017} are attractive ones. They treat video summarization as a sequence-to-sequence learning problem, where the input is the original video frame sequence and the output is the keyframe/keyshot sequence. To obtain a good video summarization, the complex and heterogeneous inter-dependency should be well considered. Both studies explore the encoder-decoder framework with \emph{Long Short-Term Memory} (LSTM) technique to model the variable-range dependencies in video summarization.  As a specific type of \emph{Recurrent Neural Network} (RNN), LSTM has shown its effectiveness in modeling long-range dependencies where the influence by the distant states on the present and future states can be adaptively adjusted and data-dependent\cite{Venugopalan2015}. Therefore, both\cite{zhang2016} and \cite{Mahasseni2017} achieve state-of-the-art performances.

However, one main drawback in such an encoder-decoder framework\cite{zhang2016}\cite{Mahasseni2017} is that it encodes all the necessary information in one single context vector no matter how long the input sequence is. Thus, the length of the intermediate code is fixed in their encoder-decoder models, which incapacitates it to give different weights to different frames in the input sequence explicitly. In this situation, all the shots/frames in the input video sequence have the same importance no matter what kind of output shots/frames are to be predicted. Due to this indiscriminate averaging of all the frames, both approaches\cite{zhang2016}\cite{Mahasseni2017} risks ignoring much of the temporal structure underlying the video. For example, considering summarizing a video ``leave home to walk dog and then come back". Since the video frames related to the ``home scene" are visually similar, it is hard for both approaches to tell the order of appearances from the collapsed vectors.

To this end, we explore the attentive encoder-decoder framework to tackle this problem in video summarization. The framework employs attention mechanism in the encoder-decoder framework by conditioning the generative process in the decoder on the encoder hidden states, rather than on one single context vector only. We name this framework \emph{Attentive encoder-decoder networks for Video Summarization} (AVS). In specific, we use attention mechanism \cite{Bahdanau2015}\cite{Luong2015} in the AVS framework, which can assign importance weights to different shots/frames of the input instead of treating all the input ones equally. In this way, it provides the inherent relations between the input video sequence and the output keyshots. Figure 1 shows an overview of AVS framework.
Compared with previous work, this paper has several essential characteristics worth being highlighted:\\
1) It proposes an \emph{Attentive encoder-decoder framework for Video Summarization}, named AVS. It is a supervised-based video summarization framework, which mimics the way of selecting the keyshots of human. To the best of our knowledge, this attentive encoder-decoder framework has not previously proposed for implementing video summarization.\\
2) It investigates the attention-based LSTM mechanism in the AVS framework, and develops two approaches to generate the video summarization. One is based on additive attention mechanism named A-AVS, the other is based on multiplicative attention mechanism, named M-AVS.\\
3) Extensive experiments are conducted on two popular video summarization datasets, including both the edited and raw video datasets. The results show the proposed M-AVS and A-AVS approaches consistently outperform the state-of-the-art ones by at least 0.8\%, 3.1\%on SumMe and TVSum datasets, respectively. These promising results verify the effectiveness of the proposed AVS framework.

The remainder of this paper is organized as follows. Section II reviews the related video summarization methods. Section III introduces the proposed AVS framework and two specific approaches. Section IV presents the experimental results and analysis. Finally, conclusions and future work are provided in Section V.
\begin{figure}[!htb]
\centering
\includegraphics[width=0.5\textwidth]{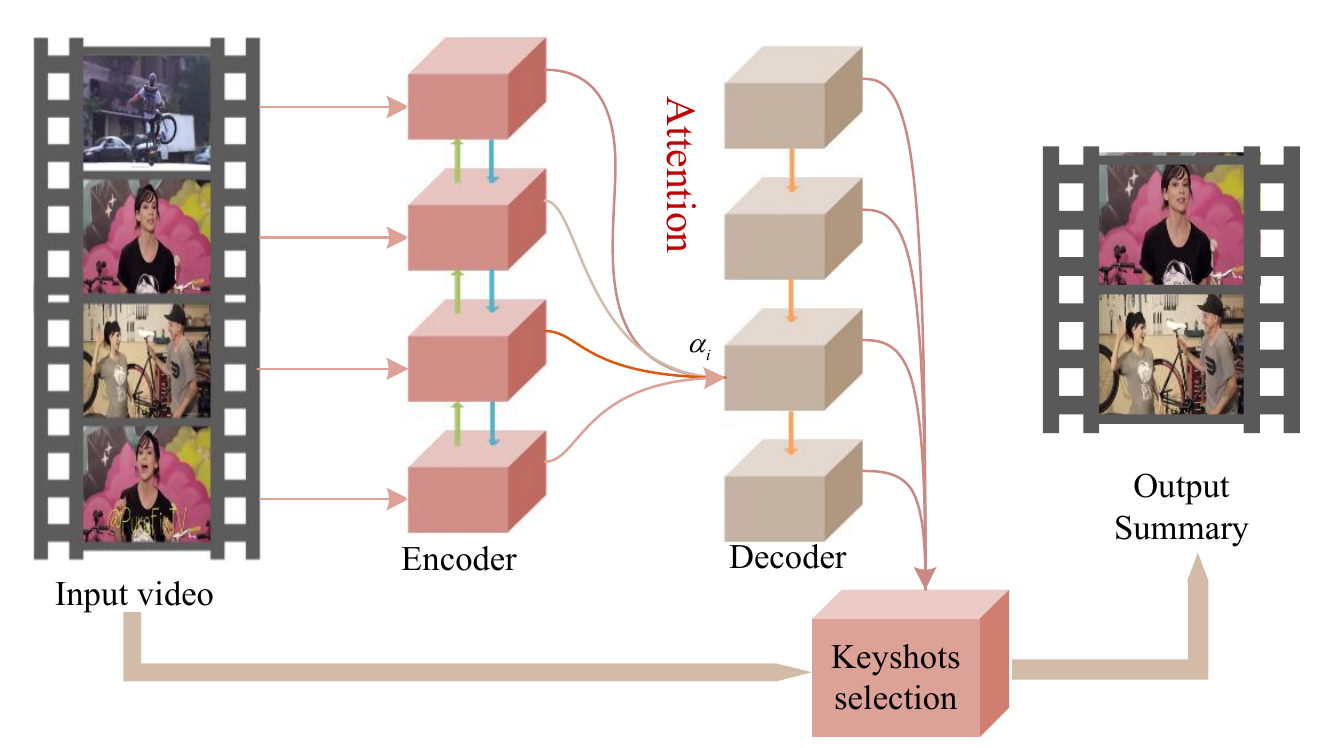}
\caption{An overview of the proposed AVS framework. It includes an encoder-decoder model and a keyshot selection model. The encoder first reads the sequence of frames and then the attention based decoder generates a sequence of importance scores. Finally, the keyshot selection model generates the keyshots based on the visual sequence and the output of the decoder.}
\label{fig:1}
\end{figure}
\section{Related Work}
According to the number of videos to be summarized, there are \emph{Single-Video Summarization }(SVS) and \emph{Multi-Video Summarization} (MVS). Specifically, SVS aims at digesting one individually long video \cite{Avila2011,Kuanar2013,Mei2015,Cong2012}, while MVS aims at summarizing a large number of short videos obtained by a query to web videos \cite{Wang2012}\cite{Ji2017}\cite{Sigurdsson2016}\cite{Nie2016}. MVS may also called query-based video summarization, and its processing method is generally different from that of SVS since it has to handle diverse query-based videos and can take advantage of the ¡°query¡± information. Furthermore, there is a direction of studying multi-view video summarization, which mainly used in surveillance scenarios to compact the videos captured from different cameras\cite{Fu2010}. In our work, we focus on SVS.

From the perspective of the learning model, there are unsupervised and supervised video summarization approaches. In the following, we will introduce their related work in detail. In particular, our work is a supervised approach. Additionally, since our work applies attention-based LSTM network and LSTM is a special type of RNN, we will further review the existing RNN-based and attention-based video summarization approaches, respectively.

\subsection{Unsupervised and Supervised Video Summarization}
Unsupervised approaches dominate the field of video summarization for a long time. They are generally designed to make the summarization meets the desired properties, such as conciseness, representativeness, and informativeness. Thus, the corresponding selection criteria for summaries include content frequency\cite{Avila2011}\cite{Kuanar2013}, coverage \cite{Mei2015}\cite{Cong2012}, relevance \cite{Wang2012}\cite{potapov2014}\cite{song2015}, and user's attention\cite{Ma2002}\cite{Ejaz2013}, etc. According to these different criteria, numerous approaches have been developed. Among them, clustering-based methods are the most popular ones \cite{Avila2011}\cite{Kuanar2013}. It clusters the visually similar frames or shots into groups, in which the group centers are considered as the representative elements of the video and therefore selected as the keyframes or keyshots. Dictionary learning is another popular technique used in unsupervised video summarization \cite{Mei2015}\cite{Cong2012}. It regards the base vectors in the dictionary model as the keyframes or keyshots since they can maximally reconstruct the visual content of the original video.

Recently, supervised video summarization approach has also received much research focus. It takes videos and their human-labeled summaries as training data to seek supervised learning methods to explicitly learn how human would summarize videos. For example, Gong \emph{et al.}\cite{Gong2014} treat video summarization as a supervised subset selection problem, and present a probabilistic model called sequential \emph{Determinatal Point Process} (seqDPP) to learn how a diverse and representative subset is selected from the training set. Potapov \emph{et al.} \cite{potapov2014} train a set of SVM classifiers to score each segment in a video with importance score, and those segments with higher scores constitute a video summary.

Besides, some work tend to directly optimize the multiple objectives for video summarization. For instance, Gygli\emph{ et al.}\cite{Gygli2015} learn to combine the criteria of representative, relevance, and uniformity to ensure the generated summaries are the most consistent with the reference ones. Specifically, they develop several submodular functions for these criteria and learn a linear combination of them using structured learning with a large margin formulation. Similarly, Li \emph{et al.}\cite{Li2017} design four functions for the criteria of representativeness, importance, diversity and storyness, respectively. And then, they build a score function to linearly combine the four functions with the maximum margin algorithm. Particularly, their proposed framework is general for both edited and raw video summarization. More recently, some deep architectures with RNN network for supervised video summarization have also been proposed\cite{zhang2016}\cite{Mahasseni2017}, which will be introduced in the next sub-section in detail.
\subsection{RNN-Based Video Summarization Approaches}
To the best of our knowledge, the only existing RNN-based video summarization approaches are \cite{zhang2016} and \cite{Mahasseni2017}. In \cite{zhang2016}, video summarization is considered as a structured prediction problem on sequential data, and a bidirectional LSTM is used to model the variable-range dependency in the video. The method is called vsLSTM. Specifically, its input is a sequence of video frames and its output is a binary indicator vector (being selected or not) or frame-level importance scores. To enhance the diversity, the authors further introduce \emph{Determinatal Point Process }(DPP) algorithm to vsLSTM, which is called dppLSTM. In \cite{Mahasseni2017}, an unsupervised generative adversarial learning model is presented, which is called SUM-GAN. Specifically, the generator is an autoencoder LSTM. Its goal is to select video frames and decode the obtained summarization for reconstructing the input video. In contrast, the discriminator is another LSTM network aiming at distinguishing between the original video and its reconstruction from the generator. Furthermore, the authors also extend SUM-GAN method to a supervised setting by adding a sparse regularization with the ground-truth summarization labels, the corresponding method is named SUM-GAN$_\text{sup}$. Both methods achieve the state-of-the-art performances in the field of video summarization. In this paper, we treat video summarization as a sequential encoder-decoder problem, and formulate it with an attention-based LSTM framework.

Video highlight \cite{Yang2015} and storyline \cite{Sigurdsson2016} have similar goals to video summarization, thus we also give brief views for existing methods using RNN network in both directions. Specifically, video highlight is a moment of major or special interest in a video. Yang \emph{et al.}\cite{Yang2015} cast it as an outlier detection problem where the non-highlights are considered as outliers. Then, they apply recurrent autoencoder with LSTM cells to model temporal dependencies to identify video highlights. Sigurdsson\emph{ et al}. \cite{Sigurdsson2016} propose a \emph{Skipping Recurrent Neural Network} (S-RNN) to learn a storyline from a photo stream. The goal of storyline is to learn the underlying visual appearances and temporal dynamics simultaneously when given hundreds of albums for a concept. Specifically, S-RNN skips through the photo sequences to extract the common latent stories.
\subsection{Attention-Based Video Summarization Approaches}
User¡¯s attention implies the concentration of mental powers upon a video segment \cite{Ma2002}\cite{Zhang2017}\cite{Ejaz2013}. If a video segment captures much attention of a user, it is more important and more likely to be a keyshot. Existing methods usually apply low-level features, such as motion and face to score the importance of video segments by modeling the user¡¯s attention. These scores join together to form an attention cue, and those on the curve crests are extracted as the keyshots to construct the summarization.

For example, Ma \emph{et al.} \cite{Ma2002} present a set of attention models via multiple sensory perceptions, such as motion, static, face, camera attention, and audio saliency. Then, these models are fused linearly and nonlinearly, respectively. Ejaz \emph{et al.} \cite{Ejaz2013} explore the static attention by using the image signature based saliency detection method, and model the dynamic attention with temporal gradients. Then, they combine both attention models non-linearly to build video summarization. Ngo \emph{et al.}\cite{Ngo2005} represent a video with a temporal graph of scenes, shots and sub-shots, where motion-based attention values are attached to each node. By modeling the evolution of a video through the temporal graph, the scene changes can be detected and the summary can be generated. More recently, to reduce the computational cost on computing the attention clues, Zhang \emph{et al.} \cite{Zhang2017} propose a simple but effective motion state change model by using a spatiotemporal slice to analyze the attention curve.

Although these attention modeling schemes have proved to be effective in video summarization, there are still some drawbacks. On the one hand, the attention curve is usually constructed with one or several low-level features. However, one feature cannot well reflect the user¡¯s attention, and several features cannot typically guarantee a correlation with what the user is interested in \cite{Boukadida2017}. On the other hand, due to the unsupervised characteristics, the existing attention-based approaches cannot take advantage of human guidance. In contrast, our proposed AVS framework can well utilize this guidance since it learns the attention mechanism in a supervised manner. Moreover, its deep neural network framework also guarantee it can capture the complex attention mechanism of viewers.

\section{The Proposed AVS Framework}

We formulate video summarization as a sequence-to-sequence learning problem, where the input is a sequence of video frames  , and the output is a sequence of keyshot. The flowchart of AVS framework is illustrated in Fig. 1. It consists of two components: an encoder-decoder model and a keyshot selection model. Particularly, the encoder-decoder model consists of an encoder and a decoder. It measures the importance of each frame. The key shots selection model aims at converting the frame-level importance scores into shot-level scores and generating summary with a length budget\cite{song2015}.

In this section, we first introduce the encoder network with a bidirectional LSTM, and then present the decoder network with attention mechanism, finally introduce the keyshot selection model briefly.
\begin{figure}[!htb]
\centering
\includegraphics[width=0.5\textwidth]{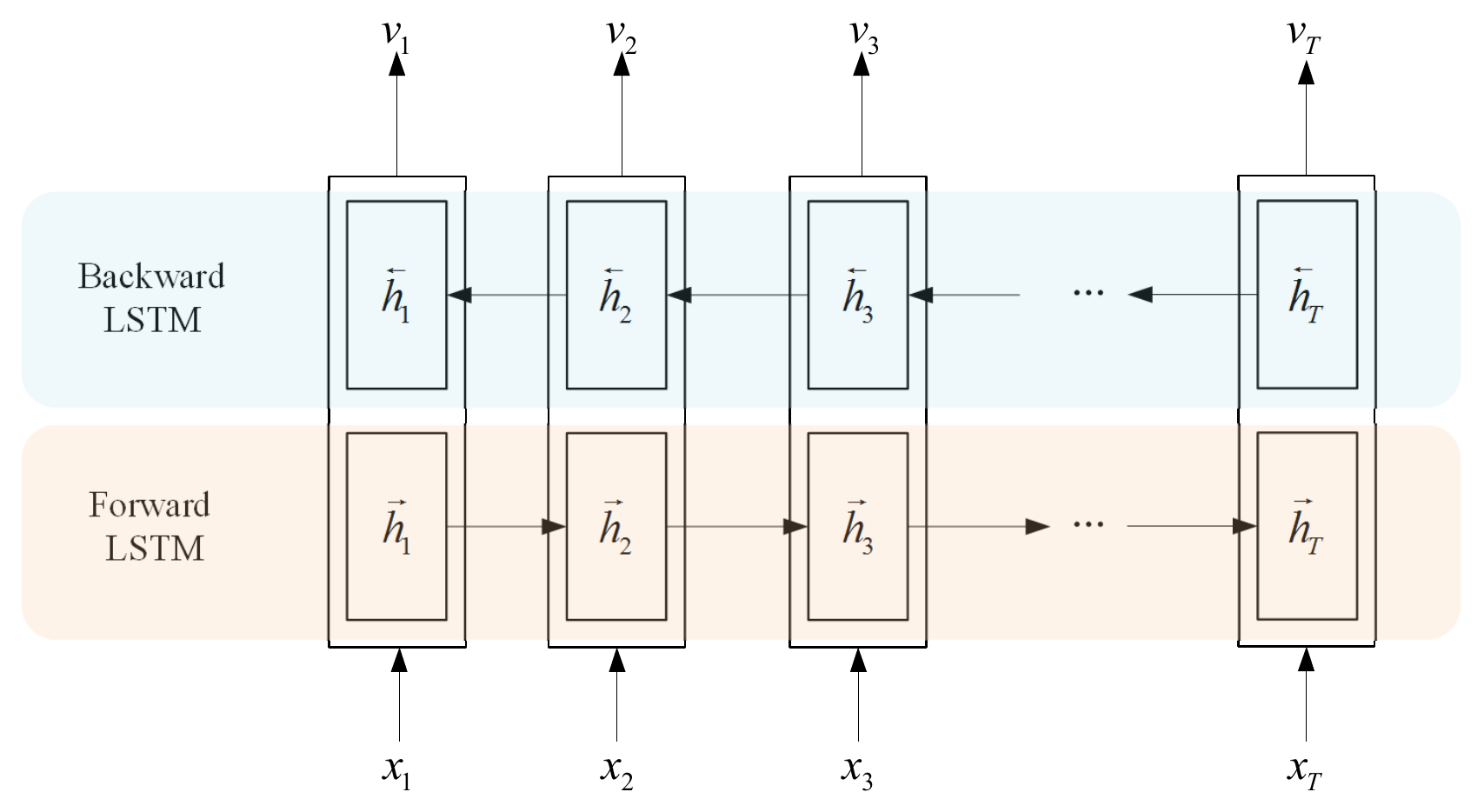}
\caption{Illustration of the BiLSTM in the proposed AVS framework.}
\label{fig:2}
\end{figure}

\subsection{Encoder with Bidirectional LSTM Network}
In a common encoder-decoder framework, an encoder converts the input sequence $X=\{x_1,x_2,...,x_T\}$ into a representation vector ${\bf{v}}{\rm{ = \{ }}{v_1}{\rm{,}}{v_2}{\rm{,}} \cdots {\rm{,}}{v_{\rm{T}}}{\rm{\} }}$.
\begin{equation}
\left[ {\begin{array}{*{20}{c}}
{{v_t}}\\
{{h_t}}
\end{array}} \right] = \phi ({x_t}),
\end{equation}where $h_t \in \mathbb{R}^n$ is a hidden state at time $t$. The architecture of an encoder $\phi$ depends on the input in a specific application. For instance, in the application of image caption \cite{Xu2015}, \emph{Convolutional Neural Network} (CNN) is a good choice. In the case of machine translation \cite{Bahdanau2015}\cite{Luong2015}, it is natural to use a RNN as the encoder, since its input is a variable-length sequence of symbols. When applied to video summarization, LSTM is the most suitable algorithm \cite{zhang2016}\cite{Mahasseni2017} since the contextual information around a specific frame is necessary for generating a video summary. It is because human relies on high-level semantic understanding of the video contents, usually after viewing the whole sequence can she/he decide which frame or shot should be selected into the summary. For example, considering summarizing a basketball game video, only a key ball that affects the game process should be selected into the summary. However, there are many goals in a basketball game, thus it is necessary to combine the scene before and after the goal to determine whether a goal is a key ball.

Inspired by outstanding performance of \emph{Bidirectional Long Short-term Memory} (BiLSTM) to encode the necessary information in a sequence\cite{Graves2005}, we select it as an encoder for taking the temporal relation of video frames into consideration. The principle of BiLSTM is to split the neurons of a regular LSTM \cite{Cong2012} into two directions, one for positive time direction (forward states), and the other for negative time direction (backward states). Moreover, those two states¡¯ outputs are not connected. By utilizing the two-time directions, the sequential information from the past and future of the current frame can be used.

The flowchart of BiLSTM is shown in the encoder part of Fig. 2. First, the forward LSTM reads the input sequence in its forward direction (from $x_1$ to $x_T$) and calculates the forward hidden states $(\overrightarrow{h}_1,\cdots, \overrightarrow{h}_T)$. Meanwhile, the backward LSTM reads the sequence in the reverse order, resulting in a sequence of backward hidden states$(\overleftarrow{h}_1,\cdots, \overleftarrow{h}_T)$. Then we obtain an annotation $v_t$ for each $x_t$ by concatenating the forward hidden state $\overrightarrow{h}_t$ and the backward one $\overleftarrow{h}_t$. That is to say, the annotation $v_t$ incorporates the information of both the preceding frames and the following frames. Due to the time tendency of an LSTM, the annotation $v_t$ can focus on the frames around $x_t$.

\subsection{Decoder with Attention Mechanism}
A decoder generates the corresponding output sequence $Y=\{ y_1,\cdots,y_m \}$ with the representation vector  from the encoder. Similar to that in the encoder, the architecture of the decoder $\psi$ is determined by the output in a specific application. In the application of video summarization, LSTM is the preferred decoder model since it runs sequentially over the output sequence \cite{Mahasseni2017}.
Generally speaking, there is a contextual relationship for each frame in a video. Due to the importance scores among frames are basically continuous in a video shot and varied among the shots, a decoder should learn the long term and short term dependency among these scores. An LSTM decoder can be written as:
\begin{equation}
\left[ {\begin{array}{*{20}{c}}
{p({y_t}|\{ {y_i}|i < t\} ,{\bf{v}})}\\
{{s_t}}
\end{array}} \right] = \psi ({s_{t - 1}},{y_{t - 1}},{\bf{v}}).
\end{equation}

However, the representation vector $\bf{v}$ in Eq. (2) is a fixed length encoding vector and cannot accurately describe the temporal characteristics of a video. To exploit the temporal ordering across the entire video, we introduce attention mechanism \cite{Bahdanau2015}\cite{Luong2015} to it. Then the decoder can be changed as:
\begin{equation}
{{\rm{V}}_t} = \sum\limits_{i = 1}^n {\alpha _t^i{v_i}}, s.t.\sum\limits_{i = 1}^n {\alpha _t^i = 1} ,
\end{equation}

\begin{equation}
\left[ {\begin{array}{*{20}{c}}
{p({{{y}}_t}|\{ {{{y}}_i}|i < t\} ,{V_t})}\\
{{s_t}}
\end{array}} \right] = \psi ({s_{t - 1}},{{{y}}_{t - 1}},{V_t}),
\end{equation}where $V_t$ stands for the attention vector at moment $t$. The attention weight $\alpha_t^i$ is a parameter to trade-off the inputs and the encoder vector. The attention mechanism allows the decoder to selectively focus on only a subset of inputs by increasing their attention weights. The attention mechanism in the LSTM decoder is shown in Fig. 3.
\begin{figure}[!htb]
\centering
\includegraphics[width=0.5\textwidth]{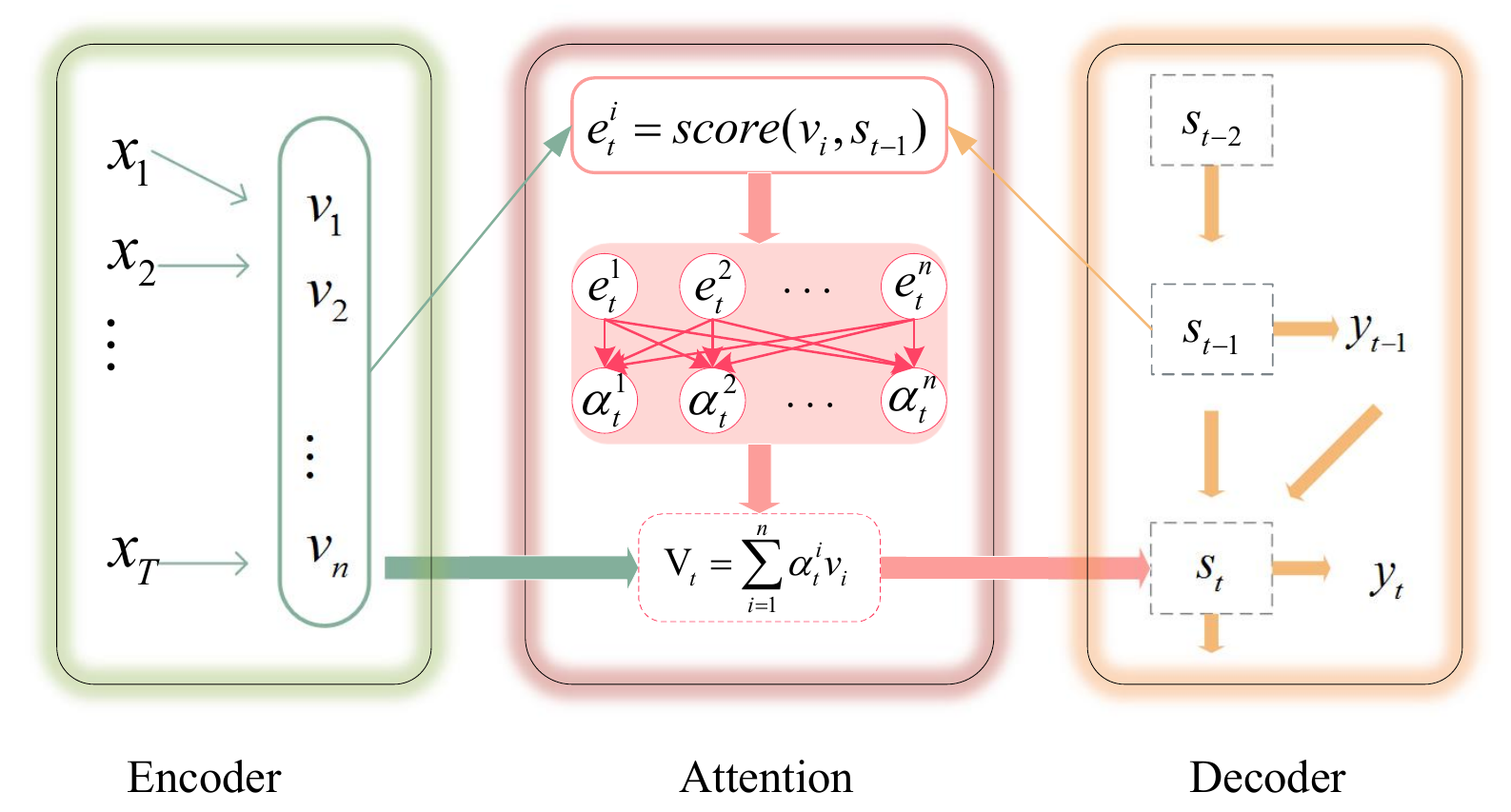}
\caption{ Illustration of the proposed attention mechanism in the LSTM decoder. To generate decoder output $y_t$ at time $t$, a score function first combines the $i$-th encoder output $v_i$ and the last hidden state of the decoder $s_{t-1}$ to obtain the relevance score $e_t^i$. Second, $e_t^i$ is normalized to gain the attention weight $\alpha_t^i$. Finally, the decoder input is obtained by weighted sum.}
\label{fig:3}
\end{figure}
The attention weight  $\alpha_t^i$  is computed at each time step $t$  , and it reflects the attention degree of the $i$-th temporal feature in the input video. To obtain $\alpha_t^i$ , the relevance score $e_t^i$  should be computed. This is because that it combines the previous hidden state $s_{t-1}$  in the LSTM decoder and the output of the encoder at time step $i$ . The score function that computes the relevance score $e_t^i$  can be written as:
\begin{equation}
e_t^i = score({s_{t - 1}},{v_i}).
\end{equation}The score function in Eq. (5) decides the relationship between the  $i$-th visual features $v_i$  and the output scores at time $t$. It can be implemented in variable ways. Concretely, we develop two models: A-AVS and M-AVS, respectively.
As shown in Fig. 4.(a), the A-AVS model applies an additive score function:
\begin{equation}
e_t^i = {w^T}\tanh ({w_a}{s_{t - 1}} + {U_a}{v_i} + {b_a}),
\end{equation}where $w,w_a,U_a$ are the weights of the additive score function and $b_a$  is the bias. These parameters are estimated together with all other parameters of the encoder and decoder networks.
The A-AVS model simply concatenates the video frames and the hidden states of the decoder. Considering a special condition that the outputs of the decoder and visual frames are matched in video summarization. That is to say, a video frame feature $v_i$  corresponds to the hidden state $s_{t-1}$  of the decoder. However, the additive function does not take full advantage of this relationship. To take a better use of the relationship between the outputs of the decoder and the visual frames, we further present an M-AVS model by exploring a multiplicative score function.
\begin{equation}
e_t^i = v_i^T{W_a}{s_{t - 1}}.
\end{equation}M-AVS model is shown in Fig. 4. (b).

Once the relevance socres $e_t^i$  for all frames $i=1,\cdots,n$  are computed, we normalize them to obtain the $\alpha_t^i$  by:
\begin{equation}
\alpha _t^i = \exp (e_t^i)/\sum\limits_{j = 1}^n {\exp (e_t^j)}.
\end{equation}

Intuitively, this implements an attention mechanism in the decoder. The decoder decides which parts of the source frames to pay attention to. Then the importance score of each frame can be computed.
\begin{figure}[!htb]
\centering
\includegraphics[width=0.4\textwidth]{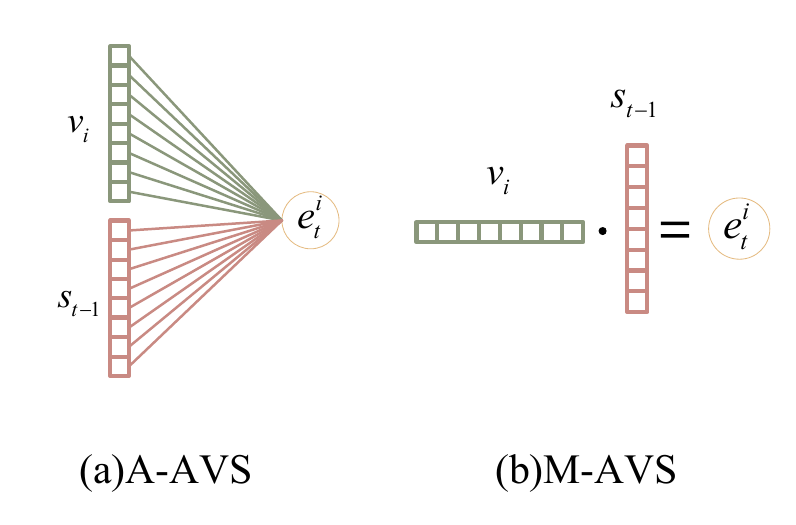}
\caption{ Illustration of the proposed score functions in the A-AVS and M-AVS model to score the relationship between the input and the output, where $v_i$ represents the $i$-th vector encoded by the encoder and $s_{t-1}$ stands for the hidden state of the decoder at time $t-1$.}
\label{fig:4}
\end{figure}

\begin{table*}
\centering
\caption{Descriptive statistics of the two datasets.}
\begin{tabular}{lcccc}
\toprule
 Dataset &\#Video &Descriptions &Duration(Min) &Annotations\\
\midrule
SumMe\cite{Gygli2014}	&25 &User generated videos of events	&1.5--6.5	&Frame-level importance scores\\
TVSum\cite{song2015}	&50	&Edited videos (10 categories)	&1--5	&Frame-level importance scores\\
\bottomrule
\hline
\end{tabular}
\end{table*}
\subsection{Keyshots Selection}
Once obtained the predicted importance scores for all frames, the remaining work is to select the keyshots to generate the video summarization. Specifically, we apply the \emph{Kernel Temporal Segmentation }(KTS) proposed by Potapov \emph{et al.}\cite{potapov2014} to segment the visually coherent frames into shots. Then it computes shot-level importance scores by taking an average of the frame importance scores within each shot. To generate keyshot-based summary, we need to solve the following optimization problem:
\begin{equation}
\max \sum\limits_{i = 1}^m {{u_i}{w_i}} {\rm{ ,}}\quad {\rm{s}}{\rm{. t}}{\rm{. }}\sum\limits_{i = 1}^m {{u_i}{l_i} \le 1,{u_i} \in \{ 0,1\} \;},
\end{equation}where $s$ is the number of shots, $w_i$ is the importance score of the $i$-th shot, and $l_i$  is the length of the $i$-th shot. Note that this is exactly the 0/1 knapsack problem, which can be solved by the dynamic programming method \cite{song2015}. The summary is then created by concatenating those shots with $u_i \neq 0$  in a chronological order.
\begin{table*}
\centering
\caption{Performance comparison (F-score) with state-of-the-art methods. Best results are denoted in \textbf{bold}.}
\begin{tabular}{lcccc}
\toprule
\text{Dataset} &Method &Feature &Supervised/unsupervised	&F-score\\	
\midrule
\multirow{9}{*}{SumMe}&
SUM-GAN$_\text{dpp}$\cite{Mahasseni2017}	&GoogleNet	&unsupervised	&39.1\\	
&Gygli \emph{et al.}\cite{Gygli2015}	&DeCAF	&supervised	&39.7\\
&Zhang \emph{et al.}\cite{zhangk2016} &AlexNet &supervised &40.9\\
&vsLSTM\cite{zhang2016}	&GoogleNet	&supervised	&37.6\\	
&dppLSTM\cite{zhang2016}	&GoogleNet	&supervised	&38.6\\	
&SUM-GAN$_\text{sup}$\cite{Mahasseni2017}	&GoogleNet	&supervised	&41.7\\	
&Li \emph{et al.}\cite{Li2017}	&VGGNet-16	&supervised	&43.1\\	
&A-AVS(ours)	&GoogleNet	&supervised	&43.9\\	
&M-AVS(ours)	&GoogleNet	&supervised	&\textbf{44.4}\\	
\midrule
\multirow{8}{*}{TVSum}&
TVSum\cite{song2015}	&HoG+GIST+SIFT	&unsupervised	&51.3\\	
&SUM-GAN$_\text{dpp}$\cite{Mahasseni2017}	&GoogleNet	&unsupervised	&51.7\\	
&vsLSTM\cite{zhang2016}	&GoogleNet	&supervised	&54.2\\	
&dppLSTM\cite{zhang2016}	&GoogleNet	&supervised	&54.7\\	
&SUM-GAN$_\text{sup}$\cite{Mahasseni2017}	&GoogleNet	&supervised	&56.3\\	
&Li \emph{et al.}\cite{Li2017}	&VGGNet-16	&supervised	&52.7\\	
&A-AVS(ours)	&GoogleNet	&supervised	&59.4\\	
&M-AVS(ours)	&GoogleNet	&supervised	&\textbf{61.0}\\	
\bottomrule
\end{tabular}
\end{table*}

\section{Experiments and Analysis}
This section first introduces the implementation details, including the datasets, evaluation metrics, and experimental settings. Then, we provide the main experimental results and parameter analysis. Next, we provide additional experiments with data argumentation. Finally, qualitative results are provided.

\subsection{Implementation Details}
\subsubsection{Datasets}

We evaluate the proposed AVS framework on two publicly available benchmark datasets: SumMe\cite{Gygli2014}, and TVSum\cite{song2015}. Most of the videos in these datasets are 1 to 10 minutes in length. Specifically, SumMe\cite{Gygli2014} consists of 25 raw videos recording a variety of events such as holidays and sports. TVSum \cite{song2015} contains 50 edited videos downloaded from YouTube in 10 categories, such as changing vehicle tire, getting vehicle unstuck, grooming an animal. The video contents in both datasets are diverse and include both ego-centric and third-person camera. In addition, both of SumMe and TVSum datasets provide frame-level importance scores for each video, which are used as the ground-truth labels. For both the two datasets, we follow the steps in \cite{zhang2016} to convert frame level scores to keyshot summaries.
Table I summarizes the key characteristics of these datasets.

\subsubsection{Evaluation Metrics}

We apply the popular F-measure as the evaluation metric \cite{Gygli2015,Gygli2014,zhang2016,Mahasseni2017,Li2017}. Similar to \cite{zhang2016} and\cite{Mahasseni2017}, our methods generate a summary $S$ which is less than 15\% in duration of the original. Given a generated summary $S$  and the ground-truth summary $G$, we compute the precision  $P$ and the recall $R$  for each pair of $S$  and $G$  based on the temporal overlaps between them, as follows:
\begin{equation}
P = \frac{\text{overlaped duration of }S \text{ and }G}{\text{duration of }S},\\
\end{equation}
\begin{equation}
R = \frac{\text{overlaped duration of }S \text{ and }G}{\text{duration of }G}.
\end{equation}
Finally, the F-measure is computed as:
\begin{equation}
F = \frac{{2 \times P \times R}}{{(P + R)}} \times 100\% .
\end{equation}

\subsubsection{Experimental Settings}
We downsample the videos into frame sequences in 2 fps. For fair comparison with \cite{zhang2016} and \cite{Mahasseni2017}, we choose to use the output of pool5 layer of the GoogLeNet \cite{googlenet2015} (1024 dimensionality), trained on ImageNet, as the visual feature for each video frame. Both proposed models have three LSTM layers, and each layer contains 256 units. The attention scale of the decoder is set as 9. As for the training/testing data, we apply the same standard supervised learning setting as \cite{zhang2016}\cite{Mahasseni2017} where the training and testing are from the disjoint part of the same dataset. We randomly leave 20\% for testing and the remaining 80\% for training.

To learn parameters in the LSTM layers, we use annotations in the forms of the frame-level importance scores. For both A-AVS and M-AVS, we stop training after 5 consecutive epochs with descending summarization F-score. The network is trained using gradient descent with a learning rate 0.15. We set attention scales to 9 and the mini-batch size to 16. For fair comparison, we run both A-AVS and M-AVS for 5 times and report the average performance.

\subsection{Comparison and Analysis}
\subsubsection{Comparison with State-of-the-art Approaches}
 Eight state-of-the-art video summarization approaches are selected for comparison with our AVS framework, including both unsupervised and supervised approaches. The performance results of the selected approaches are all from the original papers. Particularly, we are interested in comparing our performance in contrast with prior supervised approaches within the deep encoder-decoder framework, i.e., vsLSTM\cite{zhang2016}, dppLSTM\cite{zhang2016}, and SUM-GAN$_\text{sup}$\cite{Mahasseni2017}. We also choose three additional supervised approaches for comparison.The first one is Li \emph{et al.}\cite{Li2017}, which is a general framework designed for both edited and raw videos with the idea of property-weight learning. The second one is Gygli \emph{et al.}\cite{Gygli2015}, which learns submodular mixtures of objectives for different criteria directly. The third one is Zhang \emph{et al.} \cite{zhangk2016}, which learns nonparametrically to transfer summary structures from training videos to test ones. Moreover, two unsupervised approaches, SUM-GAN$_\text{dpp}$\cite{Mahasseni2017} and TVSum\cite{song2015} are chosen for comparison.

Table II shows the comparison results. We can observe that both A-AVS and M-AVS clearly outperform all the competitors in all the datasets. Specifically, on TVSum dataset, our approaches outperform the others in at least 3 absolute points. On SumMe dataset, there are almost 1 absolute points better than the state-of-the-arts. The significant improvements on TVSum against SumMe mainly lies in the fact that the association within each category of videos in TVSum is closer than that in SumMe. Thus, it is more suitable for attention mechanism to focus on the common important part of a video, which leads to a better performance on TVSum dataset.

In addition, it can be seen that the M-AVS model performs better than the A-AVS model on the two benchmark datasets in about 0.5\%-1.6\%. This is mainly due to that the multiplicative score layer makes better use of the relationship between the hidden states of the decoder and the visual feature than the additive score one. Even A-AVS has inferior performance, it outperforms SUM-GAN$_\text{dpp}$, the prior best method with deep encoder-decoder framework, in 0.8\%, 3.1\% on SumMe and TVSum datasets, respectively. The promising results prove the effectiveness and superiority of our proposed AVS framework.

\subsubsection{Importance Evaluation of Attention Mechanism}
To better verify the effectiveness of the attention mechanism in AVS framework, we abandon the attention layer in AVS to build a baseline named LSTM-VS. Figure 5 illustrates the performance comparison. It is clear to see that AVS framework outperforms the non-attention based LSTM-VS model noticeably (6\%-10\%), which also demonstrates the effectiveness of attention mechanism.
\begin{figure}[!htb]
\centering
\includegraphics[width=0.45\textwidth]{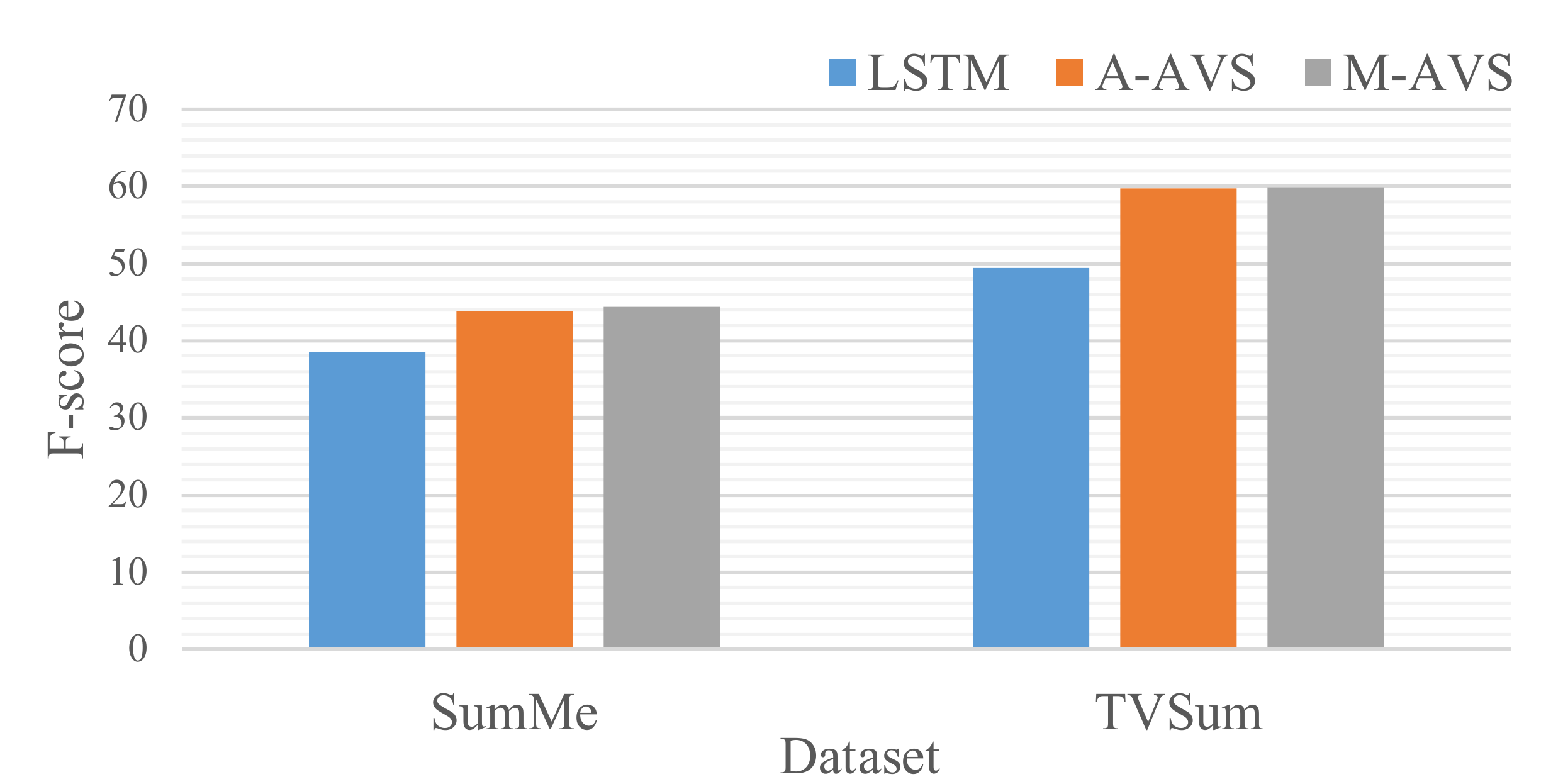}
\caption{Comparison of the proposed methods with/without attention mechanism.}
\label{fig:5}
\end{figure}

\subsubsection{Parameter Sensitive Analysis}
We evaluate the performances of our methods with different attention scales. Figure 6 shows the F-score values on two different datasets. It can be seen that the performances reach their peaks when the attention scale is around 9. It is maybe due to the fact that each shot is around 9 frames on average when we perform KTS to segment the video into shots. Therefore, we can conclude that the proposed methods will perform better when their attention scales are close to the length of shots.
\begin{figure}[!htb]
\centering
\includegraphics[width=0.45\textwidth]{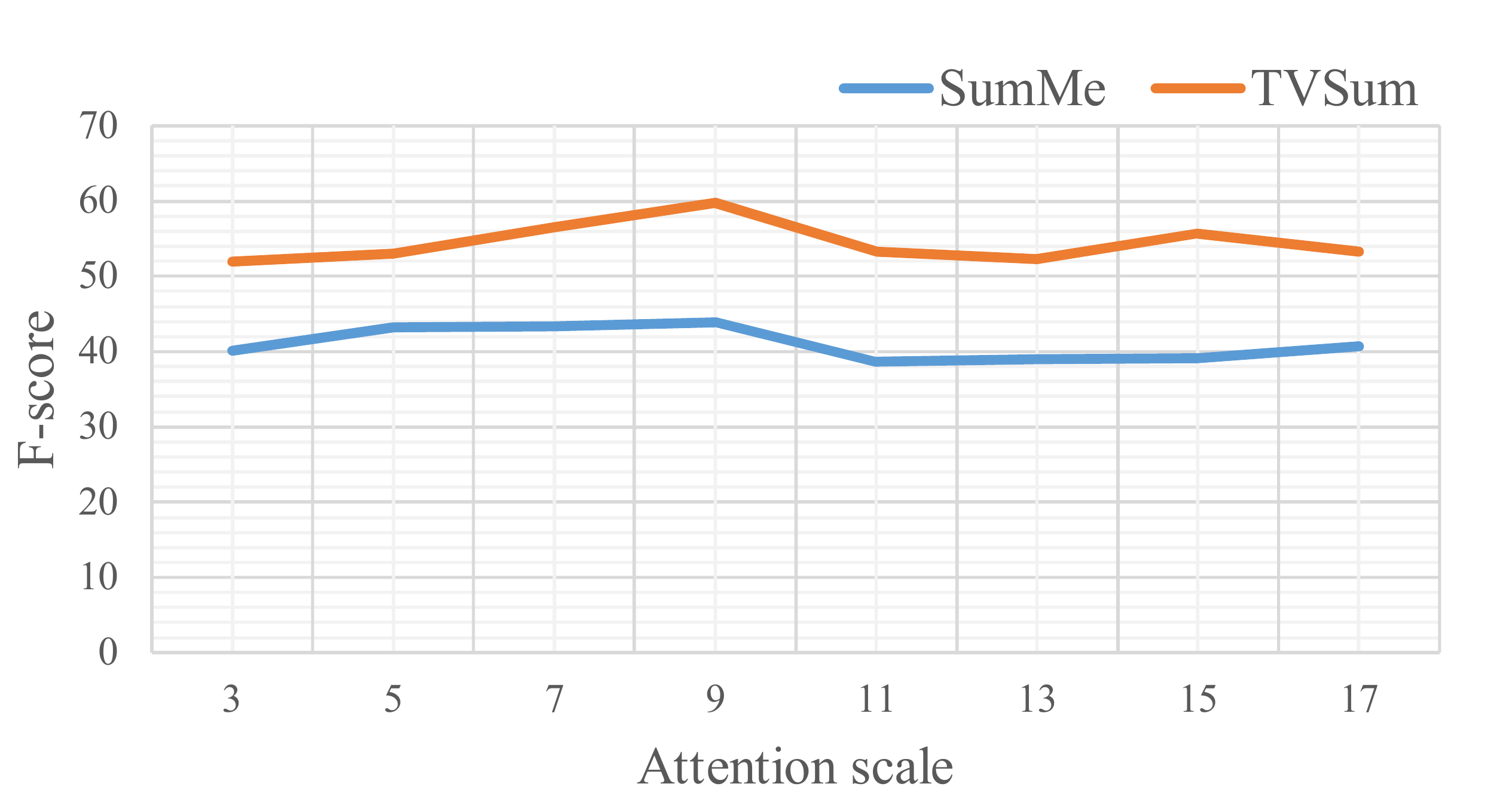}
\caption{F-score results of A-AVS model for different values of attention scales on SumMe, and TVSum datasets, respectively. }
\label{fig:6}
\end{figure}
\subsection{Augmentation Experiments}
Zhang \emph{et al}. \cite{zhang2016} and Mahasseni \emph{et al}.\cite{Mahasseni2017} augment the SumMe and TVSum datasets with OVP \cite{OVP} and YouTube\cite{Avila2011} datasets to further improve the performance on SumMe and TVSum. YouTube \cite{Avila2011} contains 50 videos selected from \emph{Open Video Project }(OVP)\cite{OVP}. The video contents include cartoons, news and sports. This dataset provides multiple user-annotated subsets of keyframes for each video, and we follow the standard approach described in \cite{zhang2016} to create a single ground truth set for evaluation. Following their settings, we implement the augmented experiments in AVS framework. Particularly, for a given dataset, we randomly leave 20\% of it for testing and augment the remaining 80\% with the other three datasets to form an augmented training dataset. The results in Table III clearly indicates that augmenting the training dataset with annotated data from other datasets improves summarization performance. For SumMe, the performances of both proposed methods rise about 0.7\%. For TVSum, the performance of A-AVS method has been improved by 1.4\%, while that of M-AVS method has been slightly improved by 0.8\%. Moreover, the augmented performances for both datasets outperform the comparative approaches. These results confirm that our models are still effective and competitive when performing data augmentation.
\begin{table}
\centering
\caption{Summarization results (F-score) with our AVS framework in the augmented setting. Best results are denoted in \textbf{bold.}}
\begin{tabular}{lccc}
\toprule
\text{Dataset} &Method &Canonical &Augmented\\	
\midrule
\multirow{4}{*}{SumMe}&
dppLSTM\cite{zhang2016}	&38.6	&42.9\\	
&SUM-GAN$_\text{sup}$\cite{Mahasseni2017}	&41.7	&43.6\\		
&A-AVS(ours)	&43.9	&44.6\\	
&M-AVS(ours)	&44.4	&\textbf{46.1}\\	
\midrule
\multirow{4}{*}{TVSum}&
dppLSTM\cite{zhang2016}	&54.7	&59.6\\	
&SUM-GAN$_\text{sup}$\cite{Mahasseni2017}	&56.3	&61.2\\		
&A-AVS(ours)	&59.4	&60.8\\	
&M-AVS(ours)	&61.0	&\textbf{61.8}\\		
\bottomrule
\end{tabular}
\end{table}
\subsection{Qualitative Results}
To better illustrate the temporal selection pattern of different variations of our approach, we demonstrate the selected frames on an example video in Fig. 7. It shows the results from vsLSTM, LSTM-VS, A-AVS, and M-AVS models on the 48-th video of the TVSum dataset. The ground-truth frame-level importance scores of the video are represented by the blue blocks. The marked orange intervals are the ones selected by vsLSTM, LSTM-VS, A-AVS, and M-AVS model respectively. We can see that the summaries generated by our methods are more uniform distribution in time than that generated by vsLSTM model. Besides, our A-AVS and M-AVS approaches select more shots with larger importance scores than the others.
\begin{figure}[!htb]
\centering
\includegraphics[width=0.45\textwidth]{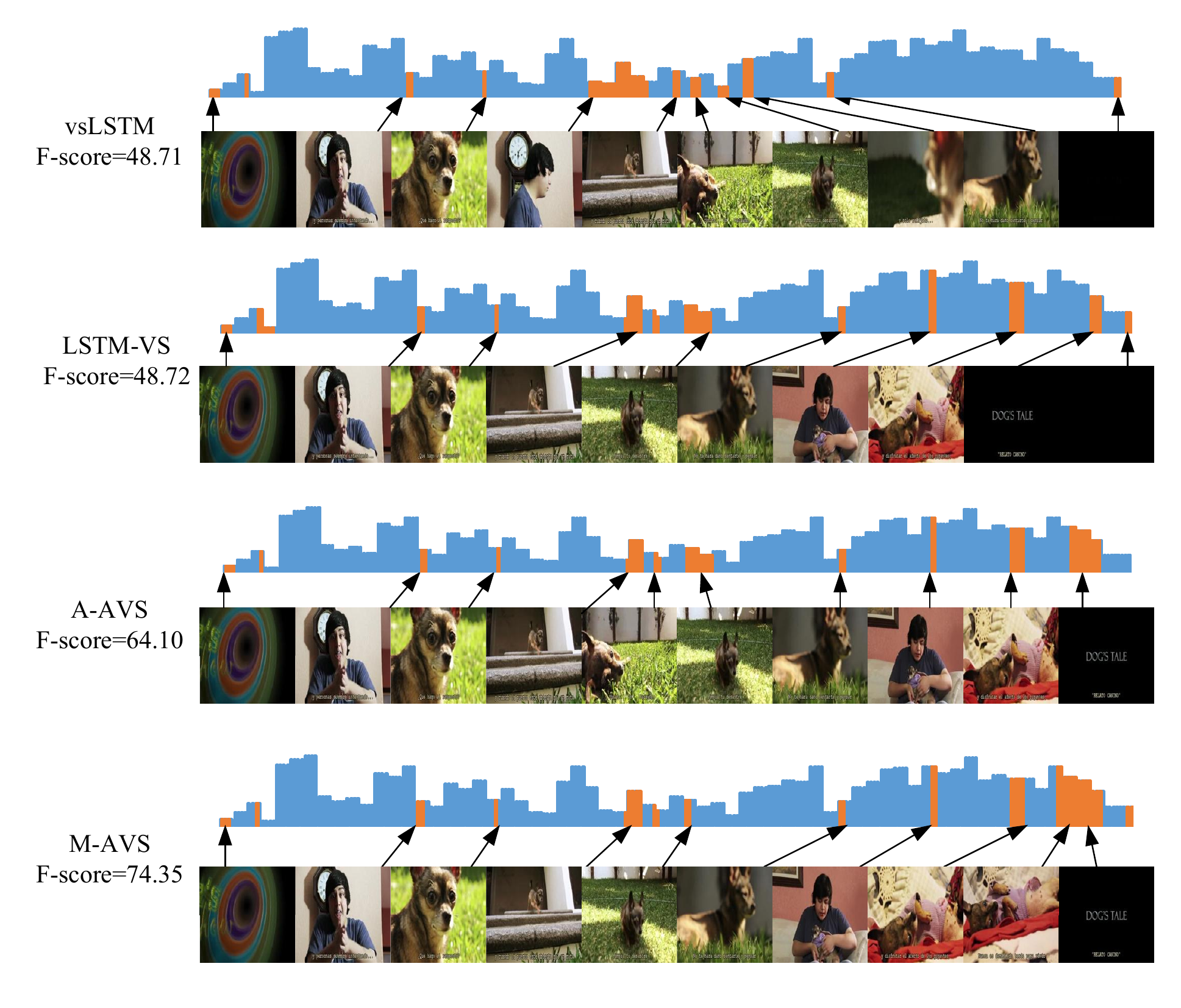}
\caption{Exemplar video summaries (orange intervals) from a sample video (the 48th video of TVSum) along with the ground-truth importance scores (blue background).}
\label{fig:7}
\end{figure}

\section{Conclusions and future work}
We propose a deep attentive framework for supervised video summarization. Specifically, two attention-based deep models named A-AVS and M-AVS are developed, respectively. To the best of our knowledge, our work is the first attempt to apply attention mechanism in deep models for video summarization. The proposed models outperform the competing methods on two benchmark datasets by 0.8\%-3\%. We also provide the qualitative analysis and parameter sensitive analysis. In addition, the augmentation experiments also verify the effectiveness and superiority of AVS framework when applied augmented data.\\
\\
In our future work, we will explore more sophisticated attention mechanism in the proposed AVS framework to obtain richer contextual information. Moreover, the existing datasets are not large enough in scale. Thus, the insufficient training data restrict the performance and development of supervised video summarization approaches. To address this problem, we will apply transfer learning \cite{zhangk2016} and \emph{Generative Adversarial Network} (GAN) \cite{Mahasseni2017} techniques to the proposed AVS framework.


%





\ifCLASSOPTIONcaptionsoff
  \newpage
\fi

\end{document}